\documentclass[conference]{IEEEtran}
\IEEEoverridecommandlockouts 

\usepackage{cite}
\usepackage[T1]{fontenc}
\usepackage[utf8]{inputenc}
\usepackage{fancyhdr}
\usepackage{ifthen}
\usepackage{caption}
\usepackage{float}
\usepackage{amsmath}
\usepackage{ wasysym }
\usepackage{ amssymb }
\usepackage{ upgreek }
\usepackage{algorithm}
\usepackage{algpseudocode}
\usepackage{pifont}
\usepackage{tikz}
\usetikzlibrary{arrows,matrix,positioning}
\usepackage{amsthm}
\usepackage{enumerate}
\usepackage{graphicx}
\usepackage{bbm}
\usepackage{wrapfig}
\usepackage{cancel}
\usepackage[bookmarks=false]{hyperref}
\usepackage{pdfpages}
\usepackage{tabularx}
\usepackage{hyperref}
\usepackage{pgfplots}
\usepackage{color}
\usepackage{mathrsfs}
\usepackage{multirow}

\newtheorem*{theorem*}{Theorem}

\newcommand{\bpg}{\begin{paragraph}{}}
\newcommand{\epg}{\end{paragraph}}
\newcommand{\setpg}[1]{\begin{paragraph}{\bf Question #1:}}
\newcommand{\unsetpg}{\end{paragraph}\newpage}
\newcommand{\bit}{\begin{itemize}}
\newcommand{\eit}{\end{itemize}}
\newcommand{\beq}{\begin{equation}}
\newcommand{\eeq}{\end{equation}}
\newcommand{\beqn}{\begin{equation*}}
\newcommand{\eeqn}{\end{equation*}}
\newcommand{\beqa}{\begin{equation}\begin{aligned}}
\newcommand{\eeqa}{\end{aligned}\end{equation}}
\newcommand{\beqna}{\begin{equation*}\begin{aligned}}
\newcommand{\eeqna}{\end{aligned}\end{equation*}}

\newcommand{\enum}{\begin{enumerate}}
\newcommand{\enuma}{\begin{enumerate}[(a)]}
\newcommand{\eenum}{\end{enumerate}}
\newcommand{\norm}[1]{||#1||}

\newcommand{\lp}{\left(}
\newcommand{\rp}{\right)}

\newcommand{\matrixcolsep}[1]{\kern#1em\vline}

\newcommand{\reals}{\mathbb{R}}

\newcommand{\bolx}{\boldsymbol{x}}
\newcommand{\boly}{\boldsymbol{y}}

\newcommand{\twonorm}[1]{\norm{#1}_2}
\newcommand{\onenorm}[1]{\norm{#1}_1}

\definecolor{dkgreen}{rgb}{0,0.6,0}
\definecolor{gray}{rgb}{0.5,0.5,0.5}
\definecolor{mauve}{rgb}{0.58,0,0.82}

\algnewcommand\algorithmicinput{\textbf{Input}:}
\algnewcommand\INPUT{\item[\algorithmicinput]}
\algnewcommand\algorithmicoutput{\textbf{Output}:}
\algnewcommand\OUTPUT{\item[\algorithmicoutput]}

\begin{document}
%
\title{Discriminative Sparsity for Sonar ATR}



\author{\IEEEauthorblockN{John D. McKay\IEEEauthorrefmark{1}, 
Raghu G. Raj\IEEEauthorrefmark{2}, 
Vishal Monga\IEEEauthorrefmark{1}, and 
Jason Isaacs\IEEEauthorrefmark{3}}
\IEEEauthorblockA{\IEEEauthorrefmark{1}
Pennsylvania State University,
University Park, PA}
\IEEEauthorblockA{\IEEEauthorrefmark{2}U.S. Naval Research Laboratory, Washington DC}
\IEEEauthorblockA{\IEEEauthorrefmark{3}Naval Surface Warfare Center Panama City, FL }
\thanks{This work is supported by the Office of Naval Research, Arlington, VA under Grant 0401531.}}

\fancypagestyle{plain}{
\fancyfoot[L]{}
\fancyfoot[C]{}
\fancyfoot[R]{\thepage}
\renewcommand{\headrulewidth}{0pt}
\renewcommand{\footrulewidth}{0pt}
}
\fancypagestyle{first}{
\fancyfoot{}
}

\pagestyle{plain}

\maketitle

\begin{abstract}
Advancements in Sonar image capture have enabled researchers to apply sophisticated object identification algorithms in order to locate targets of interest in images such as mines \cite{6825823} \cite{fandos2009sparse}.   Despite progress in this field, modern sonar automatic target recognition (ATR) approaches lack robustness to the amount of noise one would expect in real-world scenarios, the capability to handle blurring incurred from the physics of image capture, and the ability to excel with relatively few training samples.  We address these challenges by adapting modern sparsity-based techniques with dictionaries comprising of training from each class. We develop new discriminative (as opposed to generative) sparse representations which can help automatically classify targets in Sonar imaging.  Using a simulated SAS data set from the Naval Surface Warfare Center (NSWC), we obtained compelling classification rates for multi-class problems even in cases with considerable noise and sparsity in training samples.
\end{abstract}


%
\IEEEpeerreviewmaketitle

\section{Introduction}
\thispagestyle{first}

The popularity of automated underwater vehicles (AUVs) for the purposes of mine identification and location has made Sonar-specific object recognition algorithms an important topic of study.  The advantages in safety, maneuverability, and portability associated with AUVs vastly outweigh the benefits of manned vessels, but AUVs do put a particular emphasis on software that can parse non-threatening items from mines.  Indeed, a wrong classification can be extremely costly by either allowing for a mine to remain undisturbed or by causing the AUV's operator to enact expensive mine extraction measures for false alarms. 

\indent Among the many difficulties associated with Sonar automatic target recognition (ATR), we look specifically at the complications associated with noisy and blurry image classification as well as limited training sizes and foreign object detection.  Current Sonar image acquisition procedures can suffer from high noise and smearing which can significantly degrade the performance of many popular ATR algorithms.  Even for the strategies that can handle those effects, Sonar images of the same object from different angles and/or orientations can present vastly different outputs.
  This high degree of variability of Sonar images of a given target with respect to angle and orientation can
severely degrade ATR performance if not properly accounted for; and thereby puts a premium on the ability of
ATR algorithms for robustly classifying targets under low training samples (conditioned on orientation and other
relevant estimated parameters) lest one suffer from the complications arising from a large training set.
  Further, an ATR algorithm must be able to discern between that which is in its training set and foreign objects, i.e. targets that the software has not trained on.  As outlined in \cite{kriminger2015online} for the case of active Sonar ATR algorithms (which are designed to include human input), it is important for ATR approaches to include the capability to detect new objects.  
The inability to do so leads to false alarms servicing which can be very expensive in operational SonarATR
systems. The difficulty of detecting out of class samples is further accentuated by the fact that, as pointed out
above, Sonar imagery typically display high degree of translational variance (with respect to the position and
viewing angle of the sensor) which together with the effects of noise and blur tend to greatly increase the
variability within each class (which in turn is generally related to increase training set sizes). All these factors
make Sonar ATR a complicated and highly challenging problem.

\begin{figure}[t]
\centering
\includegraphics[width=4cm]{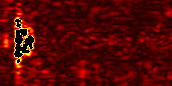}
\includegraphics[width=4cm]{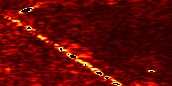}
\caption{Two SAS images of cylinders placed at different angles to a simulated AUV.  Images courtesy of RAW SAS dataset.}
\end{figure}

\indent The driving method that we employ to address some of the outstanding challenges described about is derived from the work of Wright \emph{et al} \cite{wright2009robust}.  In their seminal work, the authors demonstrated an ability to produce state-of-the-art classification rates on facial images -- despite immense noise and occlusion--with their implementation of a sparse-reconstruction based classification method (SRC), a creative adaptation of modern sparsity-based image compression concepts.  Since its publication, researchers from many different fields ranging from biology to radar have found success in applying SRC algorithms with its powerful robustness to distortions.  For Sonar, \cite{kumar2012object} demonstrated that the SRC approach can yield better accuracy results than many popular classification strategies like elastic-net or logistic regression.  In this paper we take the next step and present the application the work of Wright \emph{et al} to noisy and blurry Sonar images while also examining how well the SRC approach performed with limited training samples.  Additionally, we look at how well the sparse concentration index (SCI) metric outlined in \cite{wright2009robust} worked in discerning objects in our training set and foreign items that could be misclassified.

\indent In the next section, we present the mathematical underpinnings to the SRC approach, including the $\ell_1$ minimization technique that we chose for our experiments.  In Section 3 we demonstrate the performance of the SRC algorithm on the RAW SAS dataset obtained from the Naval Surface Warfare Center (NSWC).  To give context, we provide the results of a competing algorithm based on SIFT features and compare our SRC method to this.  Lastly, we conclude in Section 4 with a discussion of our results and directions of the future research.

\section{Methods}

In the field of compressive sensing, which looks to represent images from relatively few samples, sparsity constraints have been used with linear models to perform impressive feats of denoising and deblurring \cite{candes2008introduction}. The formulation of image reconstruction via sparsity based constraints is as follows:  let $\boly$ be the vectorized samples of noisy image measurements, $A$ an overcomplete dictionary, and $\bolx$ the coefficients vector such that $A\bolx$ represents the vectorized reconstructed image.  The idea is to minimize the number of nonzero elements in $\bolx$ while constraining the least squares error of the reconstructed image within some tolerance $\varepsilon >0$.  That is, we look to solve
\beqa\label{eq:sparseness}
\min\limits_{\bolx}\norm{\bolx}_0 \text{ subject to } \twonorm{\boly-A\bolx}<\varepsilon
\eeqa
where $\norm{\cdot}_0$ is the $\ell_0$ quasi-norm \cite{aharon2006ksvd}.  Wright \emph{et al} looked to adapt this same strategy to classification problems \cite{wright2009robust}.  In their case, the problem is to figure out which class the vector $\boly$ belonged to.  To this end, they assigned the columns of matrix $A$ to be vectorized training images corresponding to each class; i.e. if we let $A_j$ be a matrix whose columns each correspond to a vectorized training image of class $j$, then $A=[A_1\,\,A_2\,\,\dots]$.  They were then able to find the class assignment for $\boly$ by solving \eqref{eq:sparseness} and attributing the identity of the test image to the class that yields the smallest residual error.  In other words, for $\delta_j(\bolx)$ defined to be a characteristic function that outputs a vector consisting of the indexes associated with class $j$ and zeros everywhere else, they would identify the class of $\boly$ by solving
\beqna
\min\limits_{j} \twonorm{\boly-A\delta_j(\bolx)}
\eeqna 

\indent It is widely known in compressive sensing that the solution of \eqref{eq:sparseness} is a NP-hard problem.  The issue lies in the $\ell_0$ quasi-norm which renders the optimization problem to be highly non-convex and whose solution amounts to a computational intractable combinatorial search.  Because of this, it is common practice to relax the $\ell_0$ quasi-norm to its best convex approximation, the $\ell_1$ norm.  The key to this relaxation is the fact that the resulting mathematical program still promotes sparsity under suitable incoherence conditions of dictionary $A$.  Thus, instead of \eqref{eq:sparseness}, we focus on solving
\beqa\label{eq:relaxed}
\min\limits_{\bolx} \onenorm{\bolx} \text{ subject to } \twonorm{\boly-A\bolx}< \varepsilon
\eeqa
and use the same strategy as before to determine the classification.

\indent There are many different algorithms available to solve \eqref{eq:relaxed}.  \cite{yang2010fast} performed a comprehensive comparison of many different $\ell_1$ minimization approaches and found consistently that Homotopy methods worked best in denoising and classifying applications.  Based on this and other tests we conducted, we focus on Homotopy methods as our primary optimization algorithm for our SRC process.  One can find detailed descriptions of Homotopy approaches in \cite{yang2010fast} and \cite{malioutov2005homotopy}, but in short these methods use the subgradient of the objective function
\beqna
F(\bolx) = \frac{1}2 \twonorm{\boly - A\bolx} + \lambda \onenorm{x}
\eeqna
where $\lambda\in\reals$ represents a sparsity-enforcement term, and the fact that there exists a homotopy (continuous deformation) between $F(\bolx)$ and a similar objective function with a two norm relaxation of the $\ell_0$ quasi-norm as $\lambda$ approaches zero.  



\section{Experiments}

To test the SRC approach in Sonar image classification, we used the RAW SAS database obtained from the NSWC.  RAW SAS contains 13 sets of object configurations, each one with a specific arrangement of 4 objects against 11 different background settings.  While backgrounds were obtained from actual synthetic aperture Sonar measurements, the objects were realistically simulated according to their location, angles to the sensors, etc and superimposed on the scene.  Each object was snipped from its scene and saved as a target chip; this chip included possible background interference found within the rectangular window used to collect the sample.  The objects consisted of blocks, cones, cylinders, spheres, toruses, and pipes, though because the latter two were meagerly represented we tested primarily on the first four (main classes) and leave the last two, pipes and toruses (foreign classes), for our tests on foreign object detection.

\begin{table}[b]
\centering\normalsize
\begin{tabular}{|c|c|}\hline
\textbf{Object} & \textbf{Samples}\\\hline
Blocks & 88 \\
Cones & 66\\
Cylinders & 308 \\
Spheres & 66\\
Pipes & 22\\
Toruses & 22 \\\hline
\end{tabular}
\caption{Number of each object in our dataset.}\label{binsizes}
\end{table}

\indent Each of our tests had a similar experimental design.  We crafted 20 different dictionaries out of an equal number of samples from each class and tested them against 40 test images coming from 10 remaining samples per main class.  For the tests of sample size, we varied the size of the dictionaries to see how well they could classify the 40 images, while for noise and blur we kept the dictionary size constant at 100, that is, 25 from each main class.  Gaussian white noise was added at an increasing noise variance to each of the testing sets while the blurring was added at an increasing Gaussian filter. 

\indent To provide context, we also used an approach inspired by \cite{6964476} where the authors demonstrated that a SIFT-based SVM classifier can handle many of the nuances underlying Sonar image classification quite well.  For 20 trials, we took a certain number elements from each main class as our the training set and extracted their SIFT features.  We then crafted a codebook by clustering these SIFT features and found the corresponding histogram representation of all the training images.  Next, we trained an SVM based on the histogram representations and selected a testing set of 40 images (10 per main class) to apply to this classifier.  We then took each of the test images and converted all of them to their histogram representation corresponding to the codebook which ultimately allowed us to classify them with the trained SVM.  When it came to testing the performance under different training sample sizes, we varied the SIFT SVM's training equally to the SRC's.  As for noise and blur, we tested the SIFT SVM approach with a training set equal in size to the SRC, 25 samples per main class, and with a size much greater than the SRC's, 40 per main class.

\begin{figure}[t]
\centering
\begin{minipage}[t]{.2\textwidth}
\includegraphics[width=3cm]{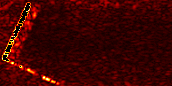}\\\\
\includegraphics[width=3cm]{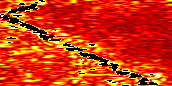}
\end{minipage}
\begin{minipage}[t]{.2\textwidth}
\includegraphics[width=3cm]{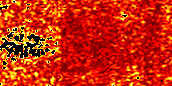}\\\\
\includegraphics[width=3cm]{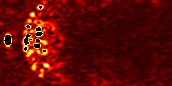}
\end{minipage}
\vspace{.3cm}
\caption{Examples of target snippets from RAW SAS; top left:  block, top right:  cone,
bottom left:  cylinder, and bottom right:  sphere}\label{examples}
\end{figure}


\subsection{Training Sample Size}





\begin{figure}[t]
\centering
\includegraphics[width=8cm]{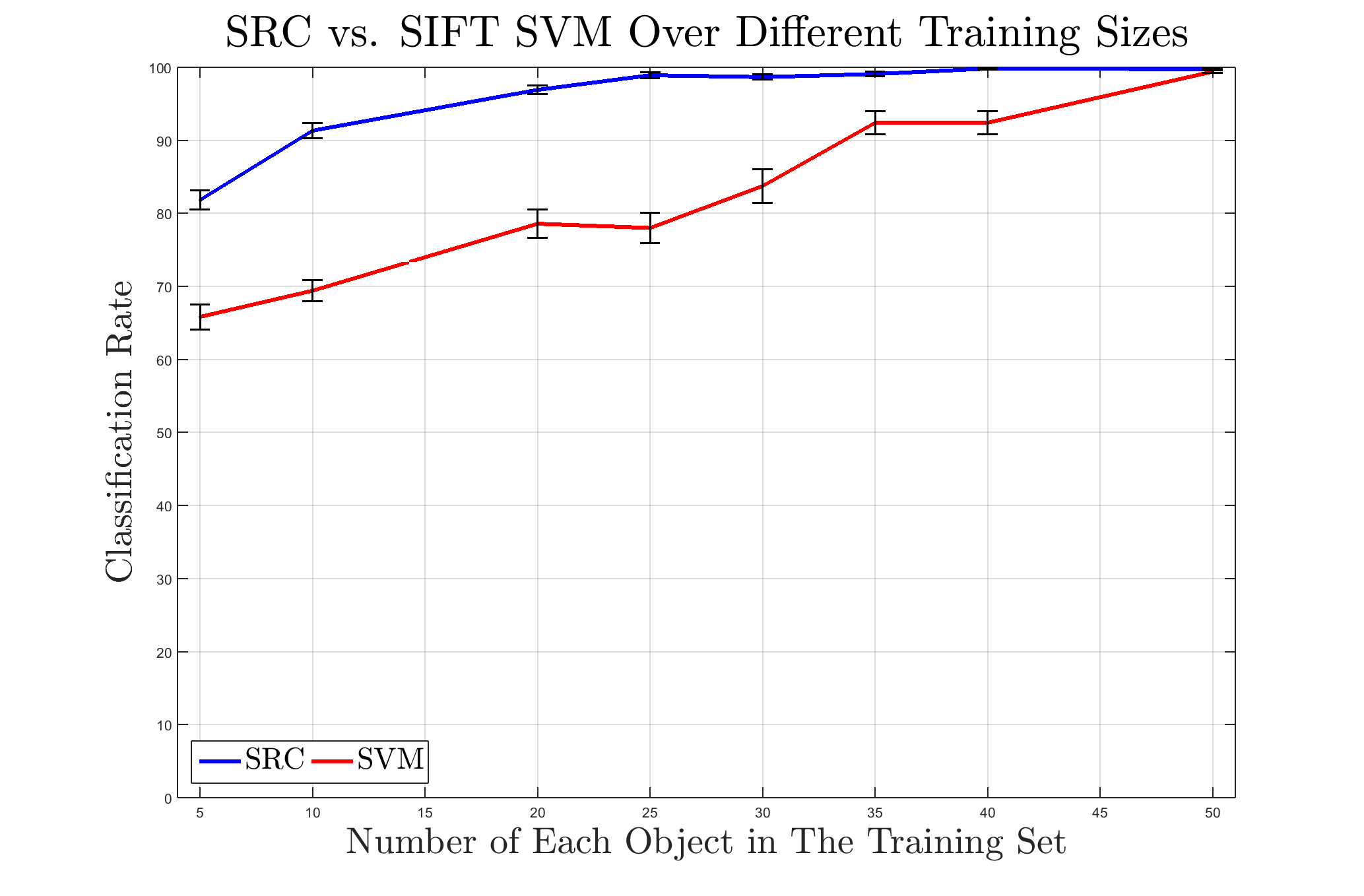}
\caption{Classification rates over various training set sizes.}\label{srcsizes}
\end{figure}

First we looked to address the performance of the SRC approach given limited training set sizes.  Figure \ref{srcsizes} provides a comparison between the SRC ans SIFT SVM approaches over various training.  Note that by 20 elements from each class - which represents less than a third of even the smallest class - our SRC dictionaries were able to obtain better than 95\% accuracy with their 40 test images.  On the other hand, the SIFT SVM approach was unable to yield an average accuracy rate above 90\% until 35 samples from each class and was vastly outperformed by the SRC method for the smaller training sizes.

\begin{figure}[H]
\centering
\includegraphics[width=4.2cm]{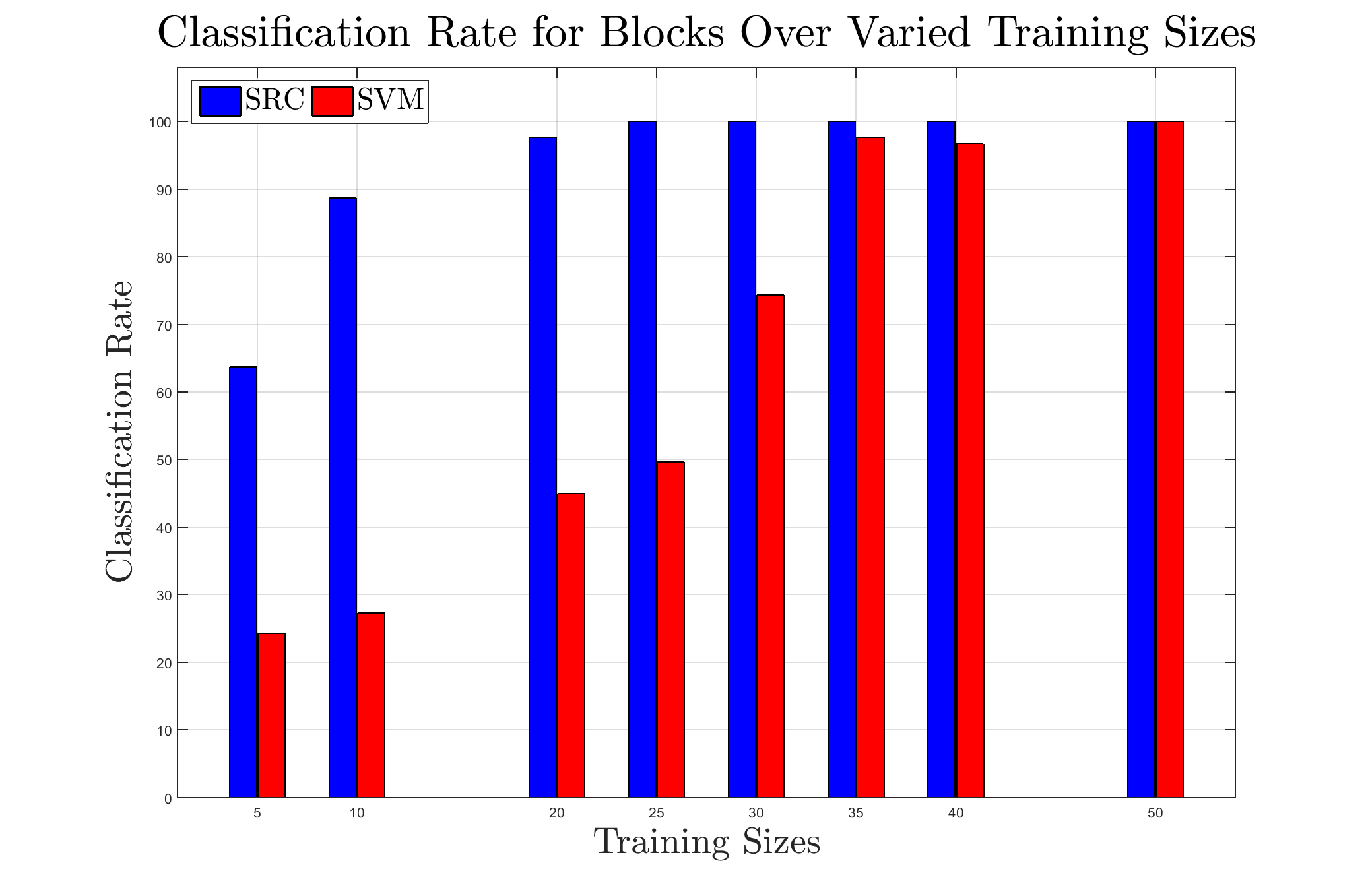}
\includegraphics[width=4.2cm]{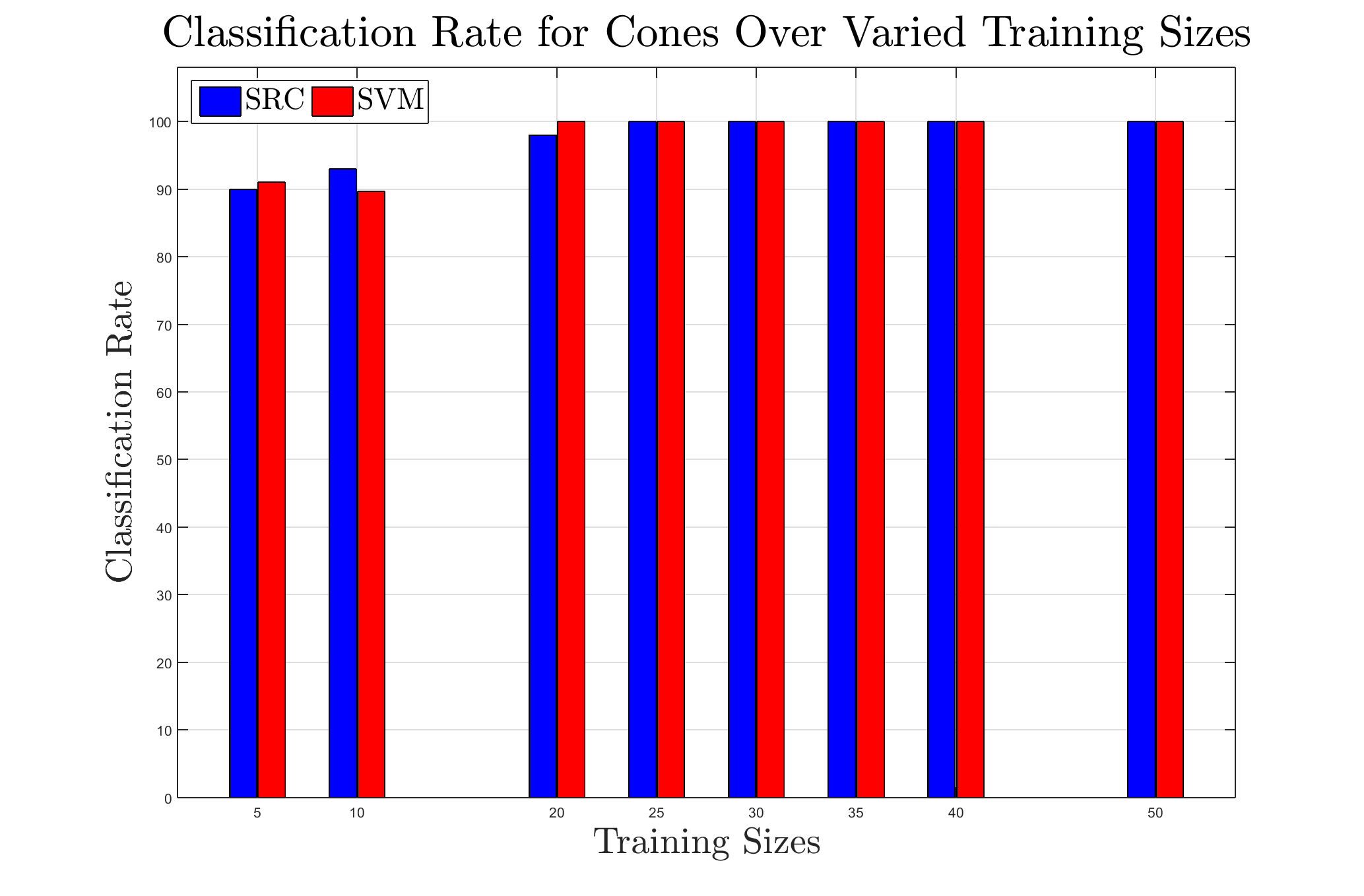}\\
\includegraphics[width=4.2cm]{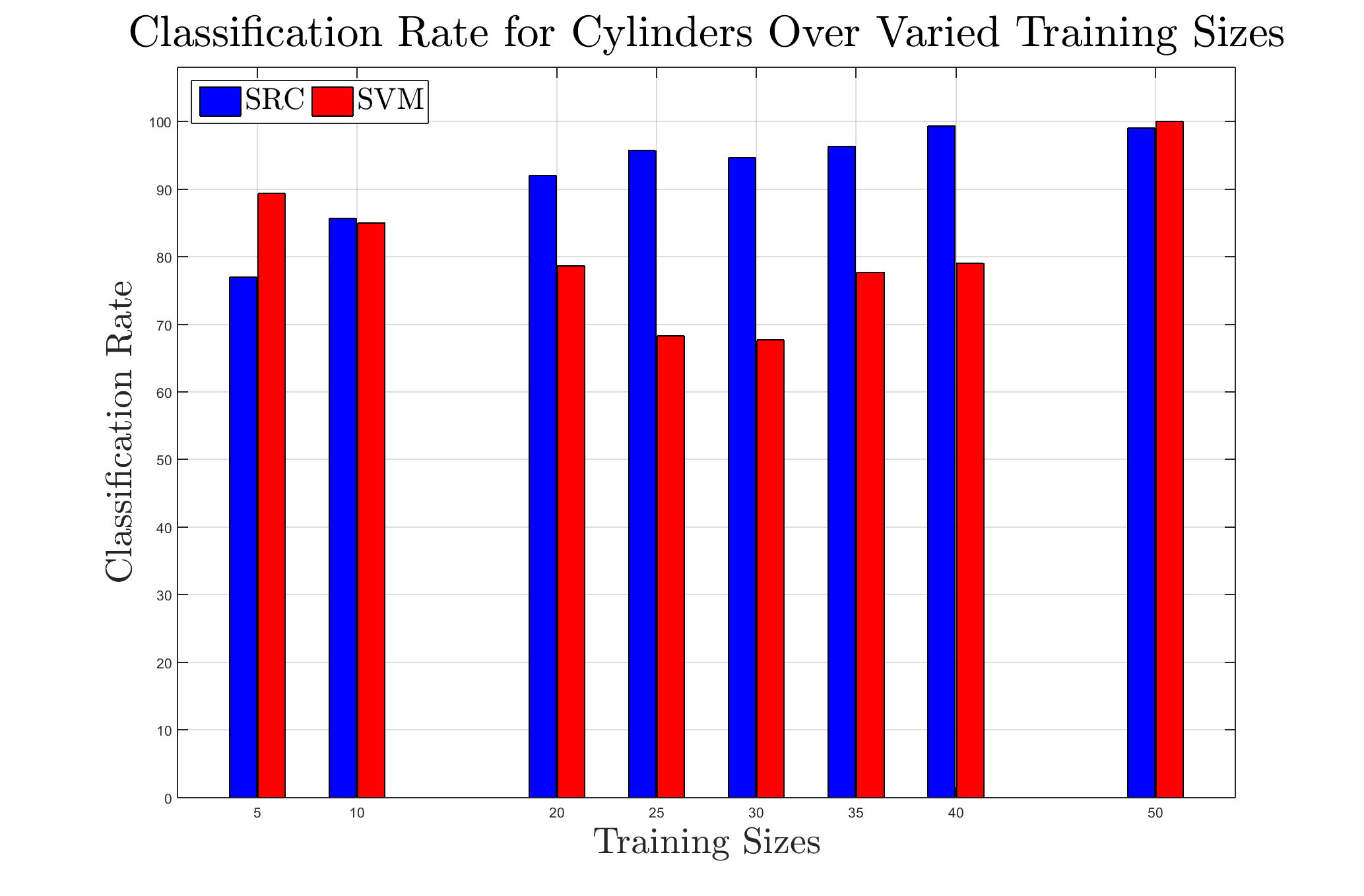}
\includegraphics[width=4.2cm]{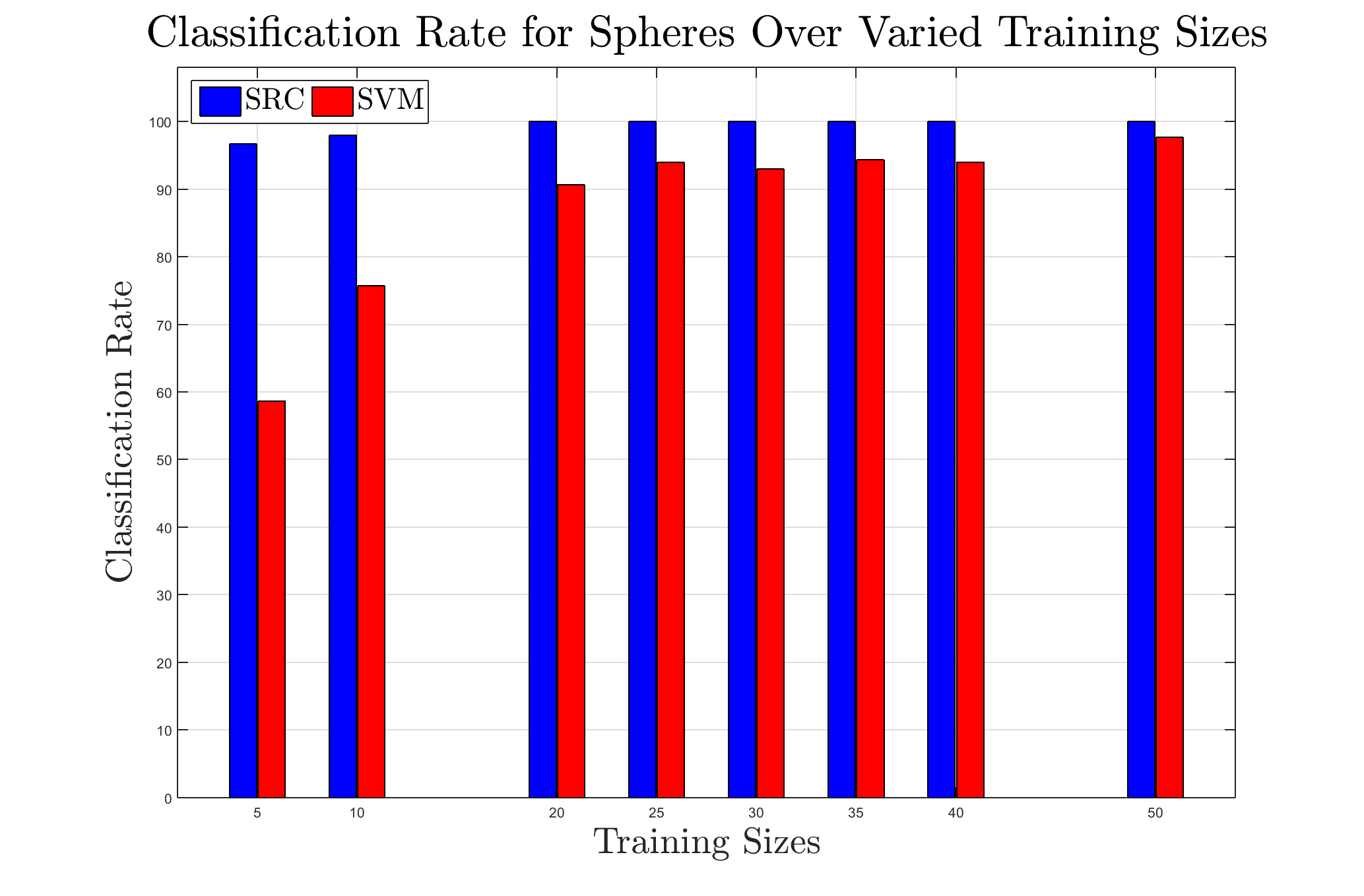}
\caption{Class specific classification rates over varied training.}\label{sizebar}
\end{figure}

\indent With regards to the classification rate of each individual classes, Figure \ref{sizebar} shows on average how well they were identified over the different training sizes.  The block and sphere cases stand out as examples wherein the SRC algorithm was able to perform at a high rate even in limited training settings -- especially the spheres where even with as few as 5 training samples, the SRC provided classification rates above 95\%.  The cones seemed to give neither algorithm much of a challenge while the cylinders proved to be an interesting case in variability.  The many different angles and orientations that we had for the cylinders gave the SIFT SVM trouble in providing predicable results, while the SRC was still able to demonstrate the monotonically increasing classification rate pattern we would expect in this experiment.  This speaks to the potential reliability the SRC algorithm may offer that Sonar ATR needs:  predictability. 

\subsection{Classification Under Noise}

Like with limited training samples, Wright \emph{et al} showed how well the SRC approach can handle noise.  In Sonar applications where a misclassification can be costly, the onus is on the software to demonstrate a resiliency to noise.  In is in this setting that we performed our next set of experiments on the RAW-SAS dataset.

\begin{figure}[t]
\centering
\includegraphics[width=8cm]{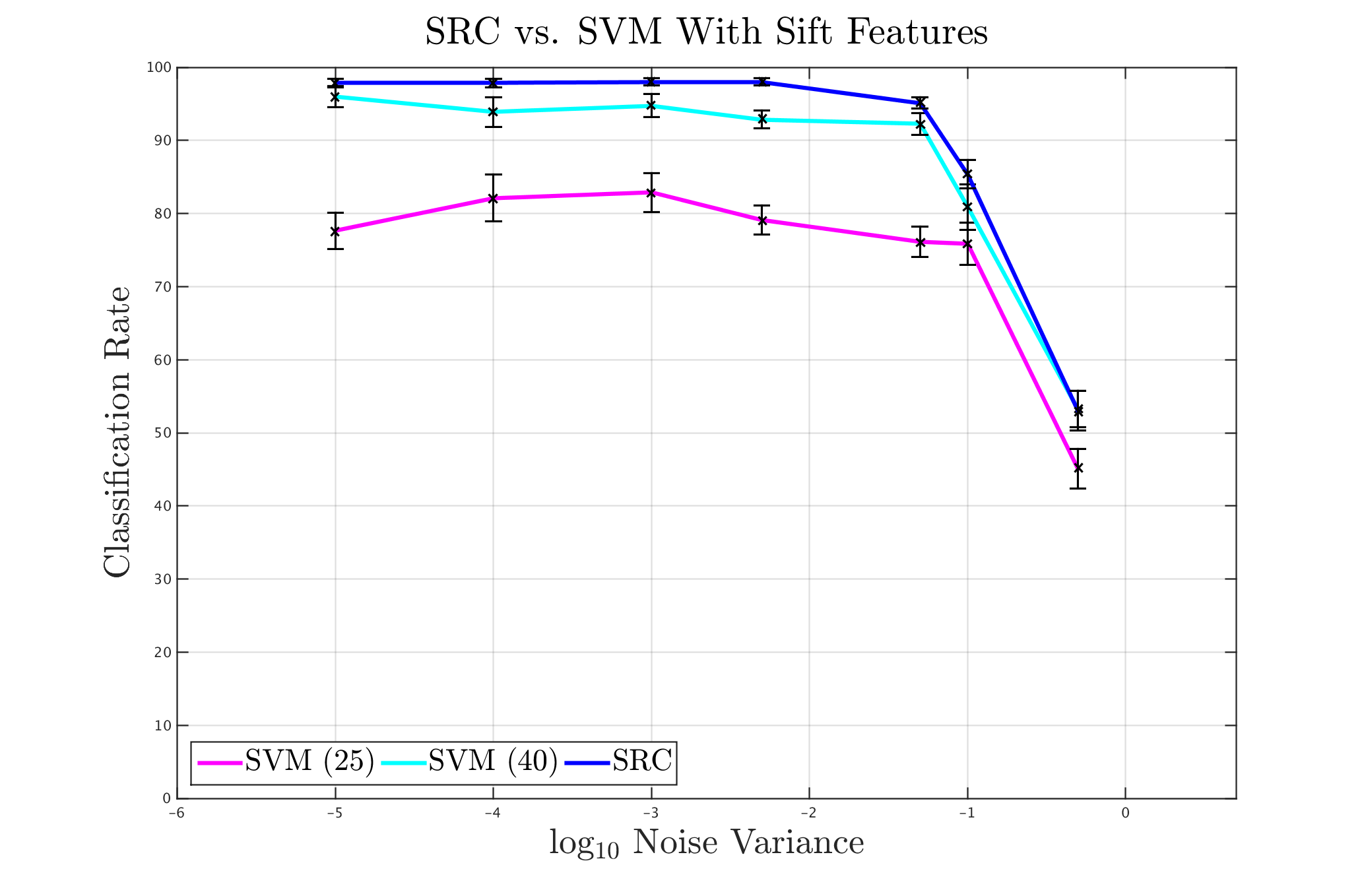}
\caption{Noisy image classification rates with standard error bars.}\label{noisecomp}
\end{figure}

\indent In every case the SRC method outperformed the SIFT SVM approach.  As demonstrated in Figure \ref{noisecomp}, the SRC approach was able to provide much higher classification rates than the SIFT SVM in equal training environments and was even able to slightly edge the SIFT SVM with many more training samples in all but the heaviest noise scenarios.  While this confirmed our suspicions given how successful Wright \emph{et al} were with their facial databases, it is still impressive how well the SRC approach handles Sonar images especially considering the presence of significant background clutter already present even without the noise.  That said, the scenes with the highest noise did give our SRC approach trouble, though not as much as the SIFT SVM with equal training.

\begin{figure}[b]
\centering
\begin{minipage}[t]{.23\textwidth}
\includegraphics[width=4cm]{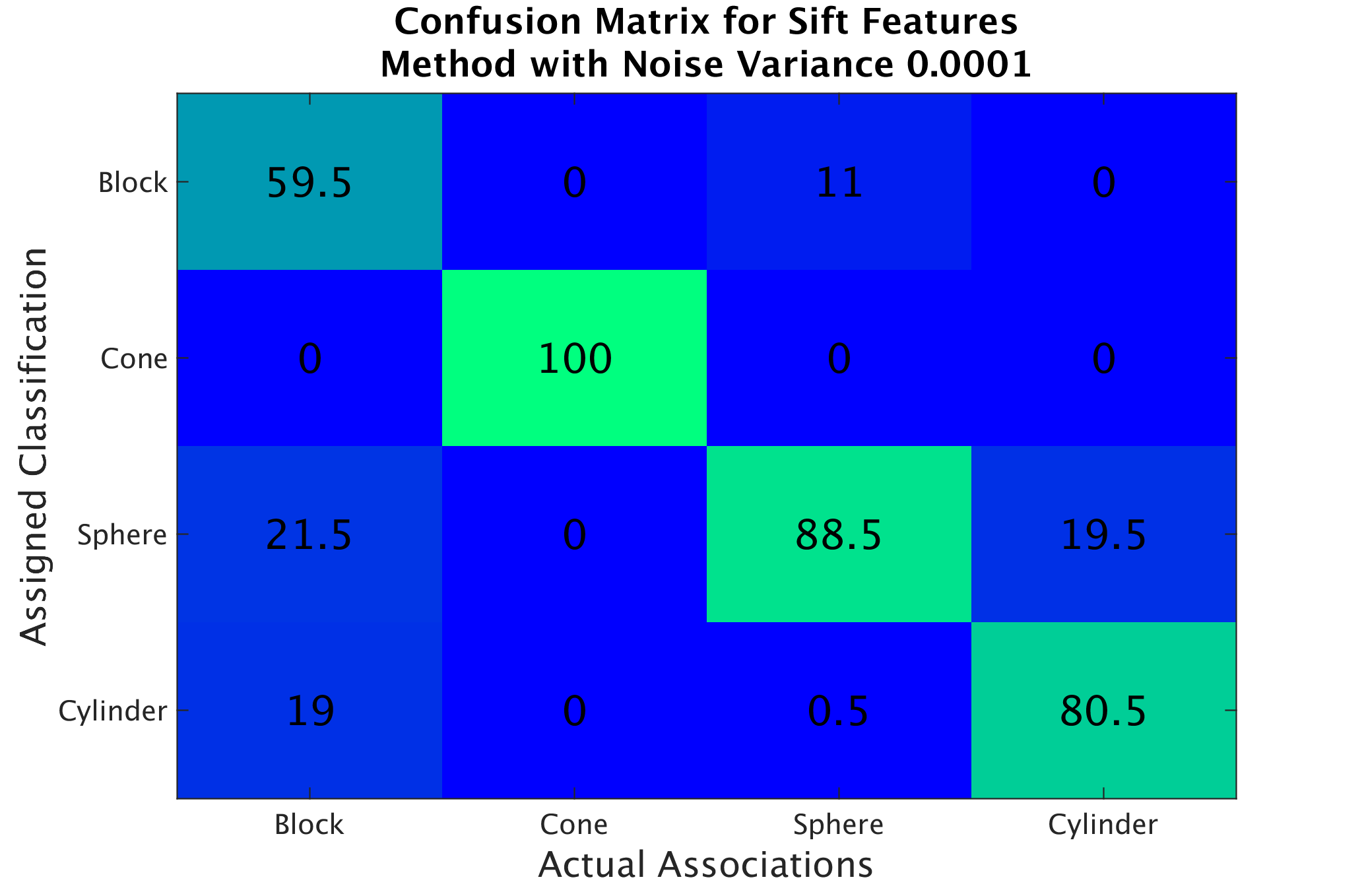}\\\\
\includegraphics[width=4cm]{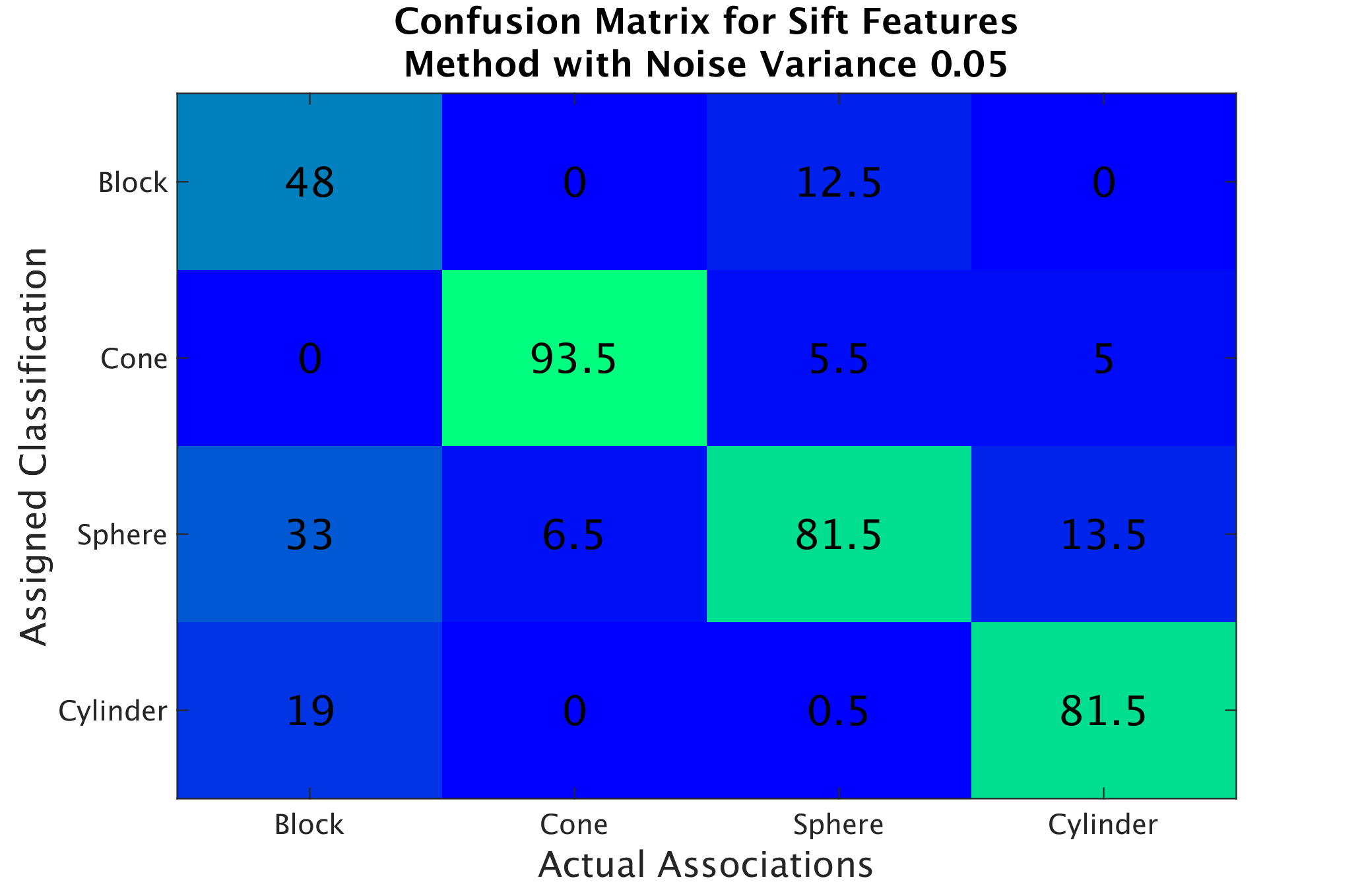}
\end{minipage}
\begin{minipage}[t]{.23\textwidth}
\includegraphics[width=4cm]{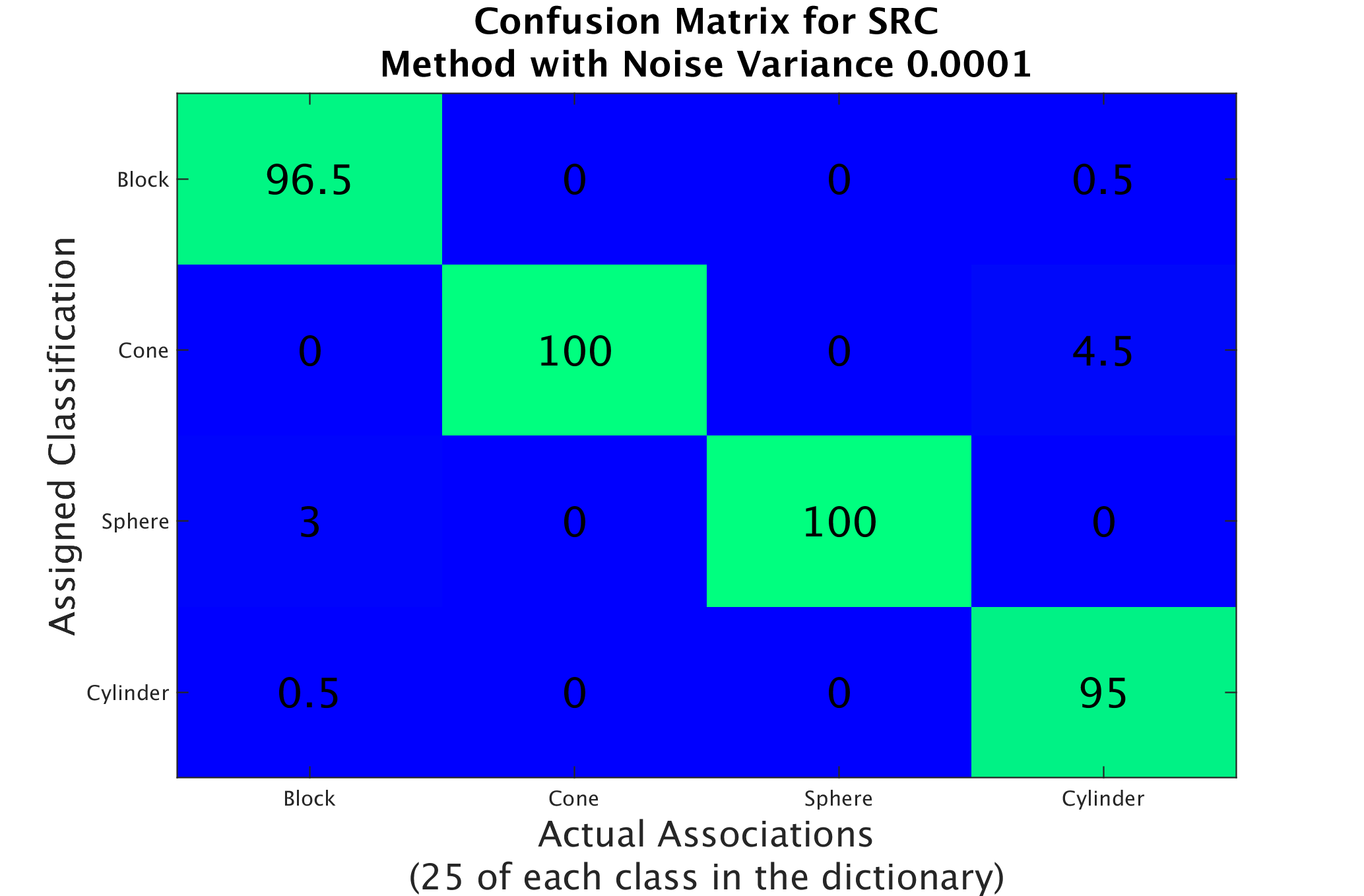}\\\\
\includegraphics[width=4cm]{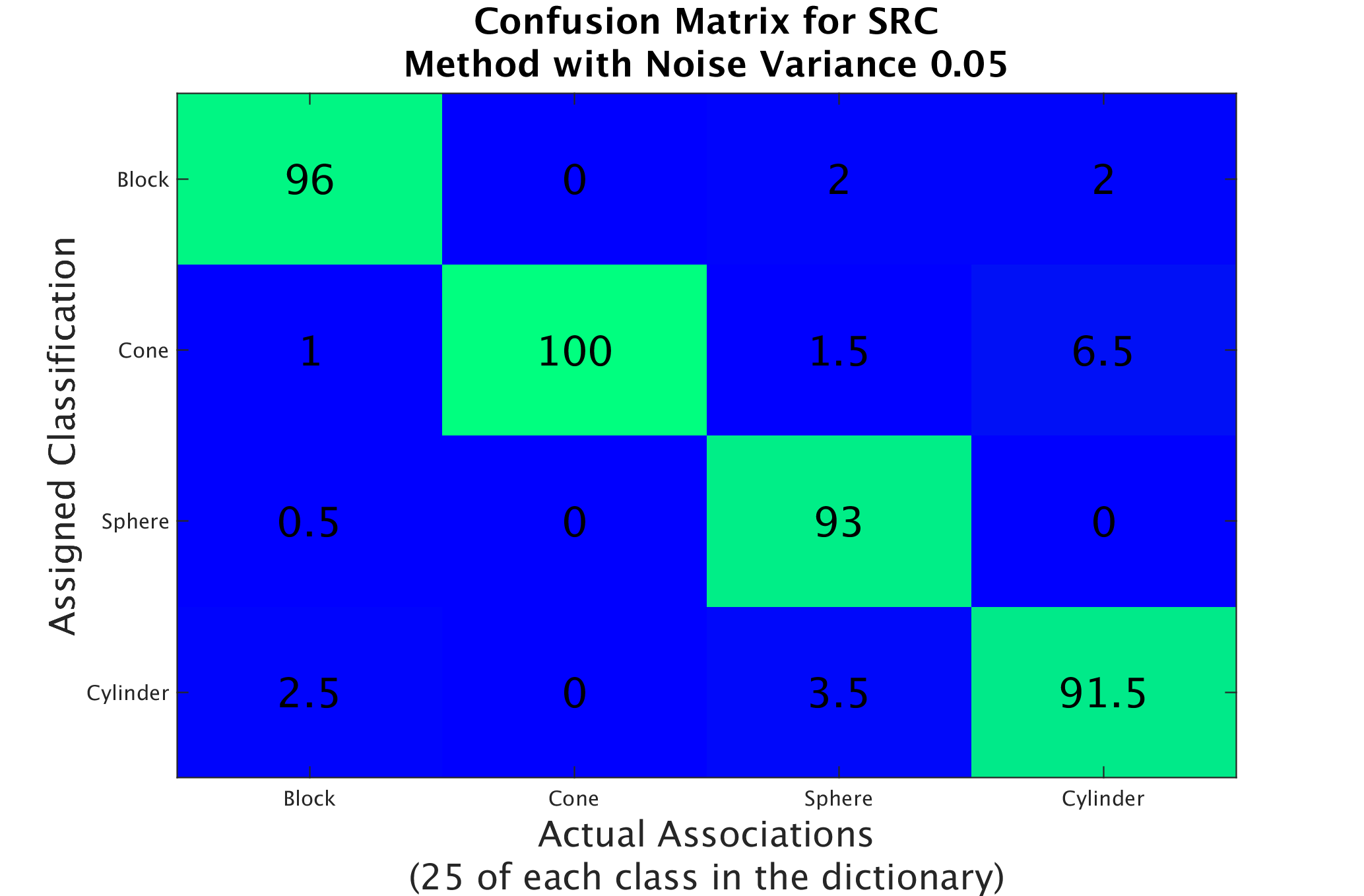}
\end{minipage}
\caption{Scaled confusion matrices for lower and higher variance noise for the SRC and SIFT SVM with equal training.  The columns represent the actual tested image class and the rows their classified class (sums of columns are 100\%).}\label{heatmaps}
\end{figure}

\indent Figure \ref{heatmaps} gives an idea of how each main class faired with the SRC and SIFT SVM approach with equal training.  These image depict how the SIFT SVM had problems deciphering between spheres and cones; as \ref{examples} shows, they do share similar geometric characteristics making for the possibility of difficultly when it came to parsing between them with SIFT features.  The SRC method, on the other hand, did not appear to have trouble with any one class and, instead, saw a degradation in classification rates overall.

\indent It is worth noting how the SRC approach was able to rightly identify images 95\% of the time or better up until the noise reached a variance of  $.1$ and images with approximate SNR values of $-10dB$.  Therefore, given the relative performance, the SRC demonstrated promise but still has trouble under heavy noise.


\subsection{Classification Under Blurring}

\begin{figure}[t]
\centering
\begin{minipage}[t]{.2\textwidth}
\centering
\includegraphics[width=3.5cm]{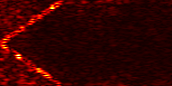}\\\vspace{.1cm}
\includegraphics[width=3.5cm]{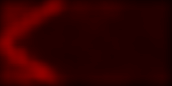}
\end{minipage}
\vspace*{.05cm}
\begin{minipage}[t]{.2\textwidth}
\centering
\includegraphics[width=3.5cm]{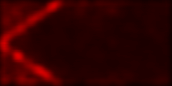}\\\vspace{.1cm}
\includegraphics[width=3.5cm]{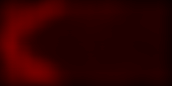}
\end{minipage}
\caption{Example blurring on a block; top left is the original, bottom left is a lower level blur, top right is a higher blur intensity, and bottom right is the highest.}\label{blurs}
\end{figure}

Much like noise, AUV systems must be ready to handle non-trivially blurred Sonar images.  We tested our SRC algorithm and the SIFT SVM approach with increasingly blurry images using simple Gaussian filters on test images.  We use the term ``blurring intensity'' to refer to the value $b\in\reals$ that represents the pixel dimension of the blurring as well as the Gaussian standard deviation for the filter.  To get an idea of how the blurring progressed in our experiment, Figure \ref{blurs} gives an example of blurred test images.

  
\begin{figure}[t]
\centering
\includegraphics[width=8cm]{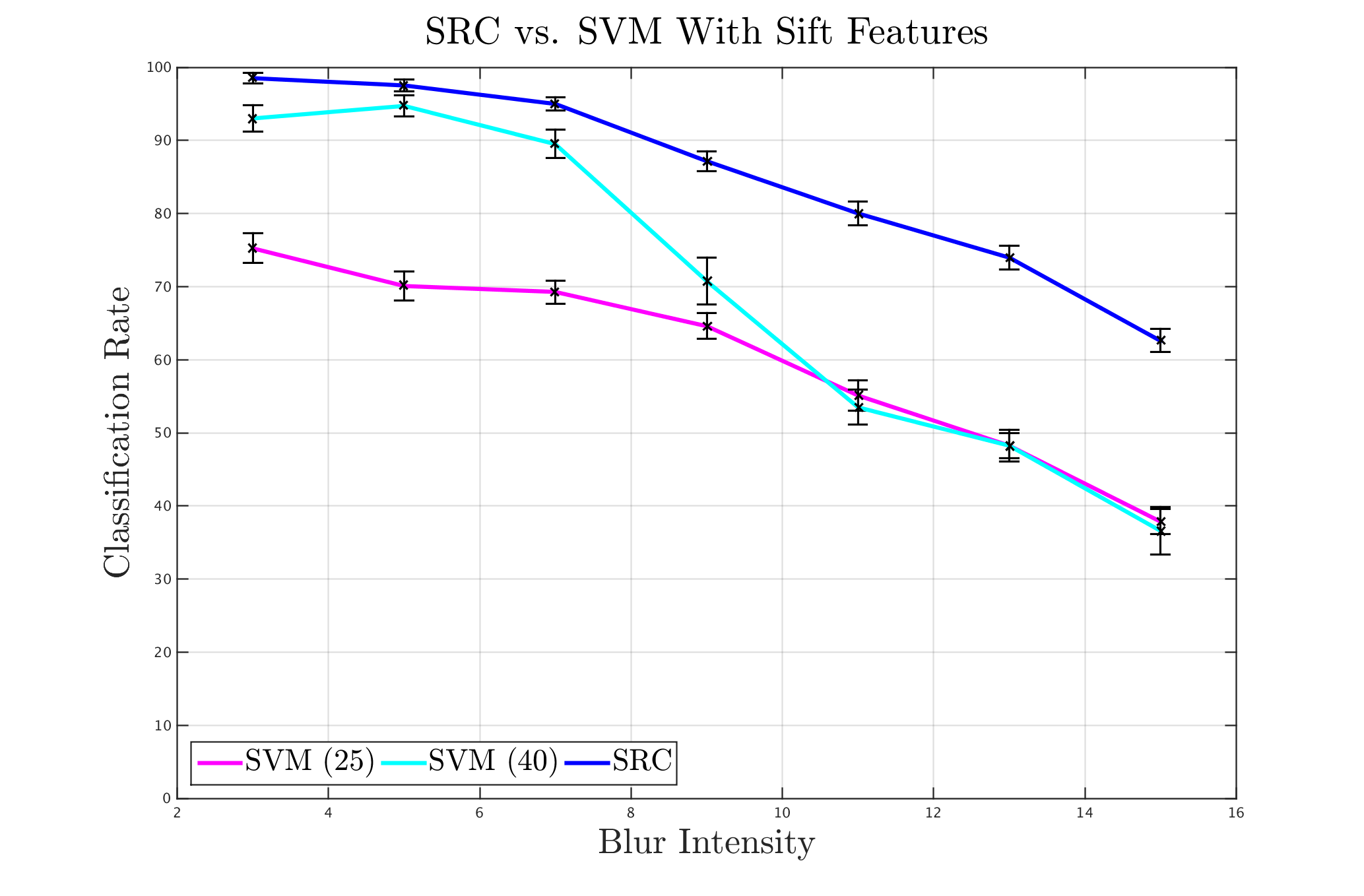}
\caption{Blurring experiment results for SRC and SIFT SVM with standard error bars.}\label{blurcomp}
\end{figure}

\begin{figure}[b]
\centering
\includegraphics[width=4.2cm]{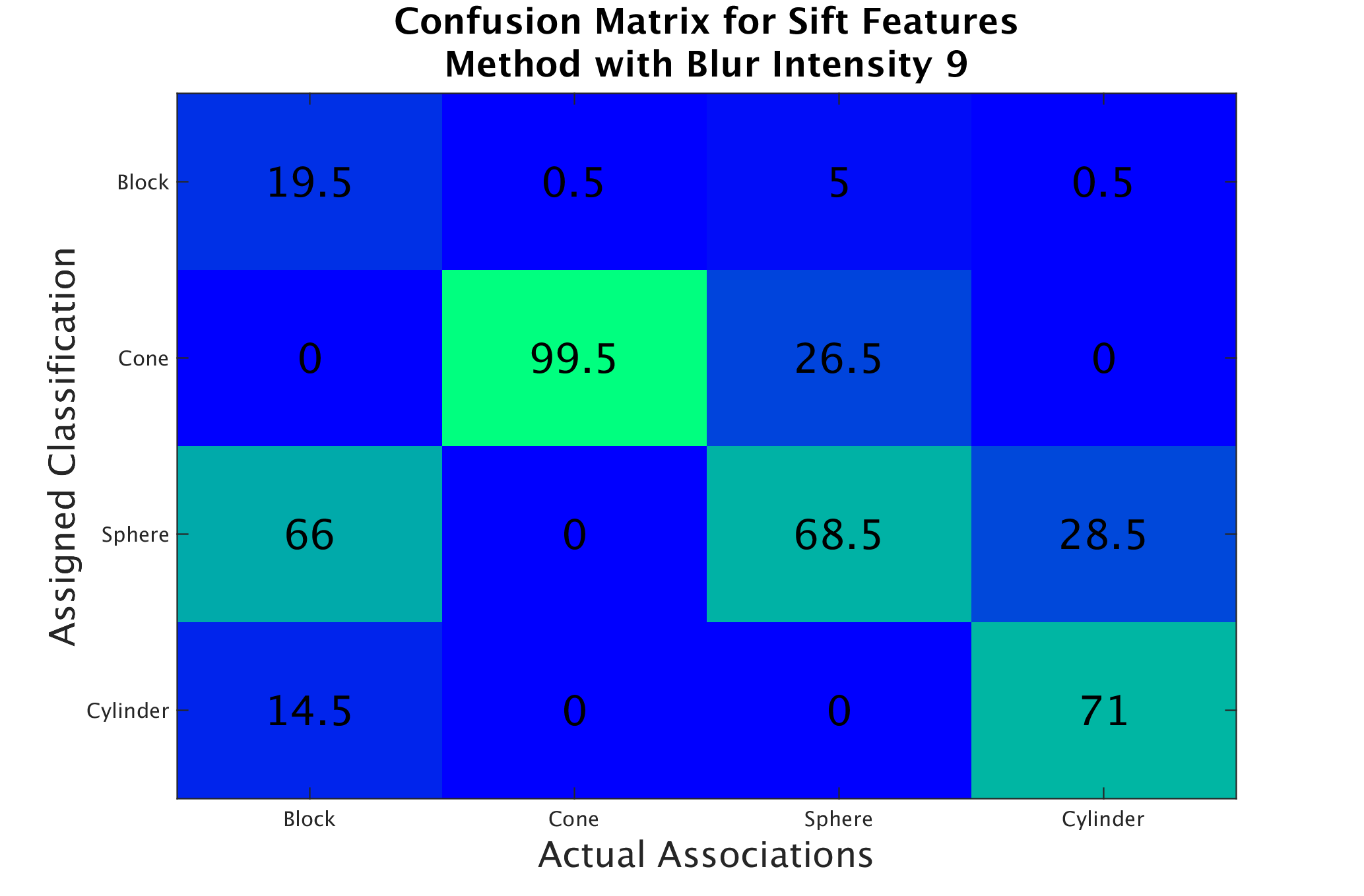}
\includegraphics[width=4.2cm]{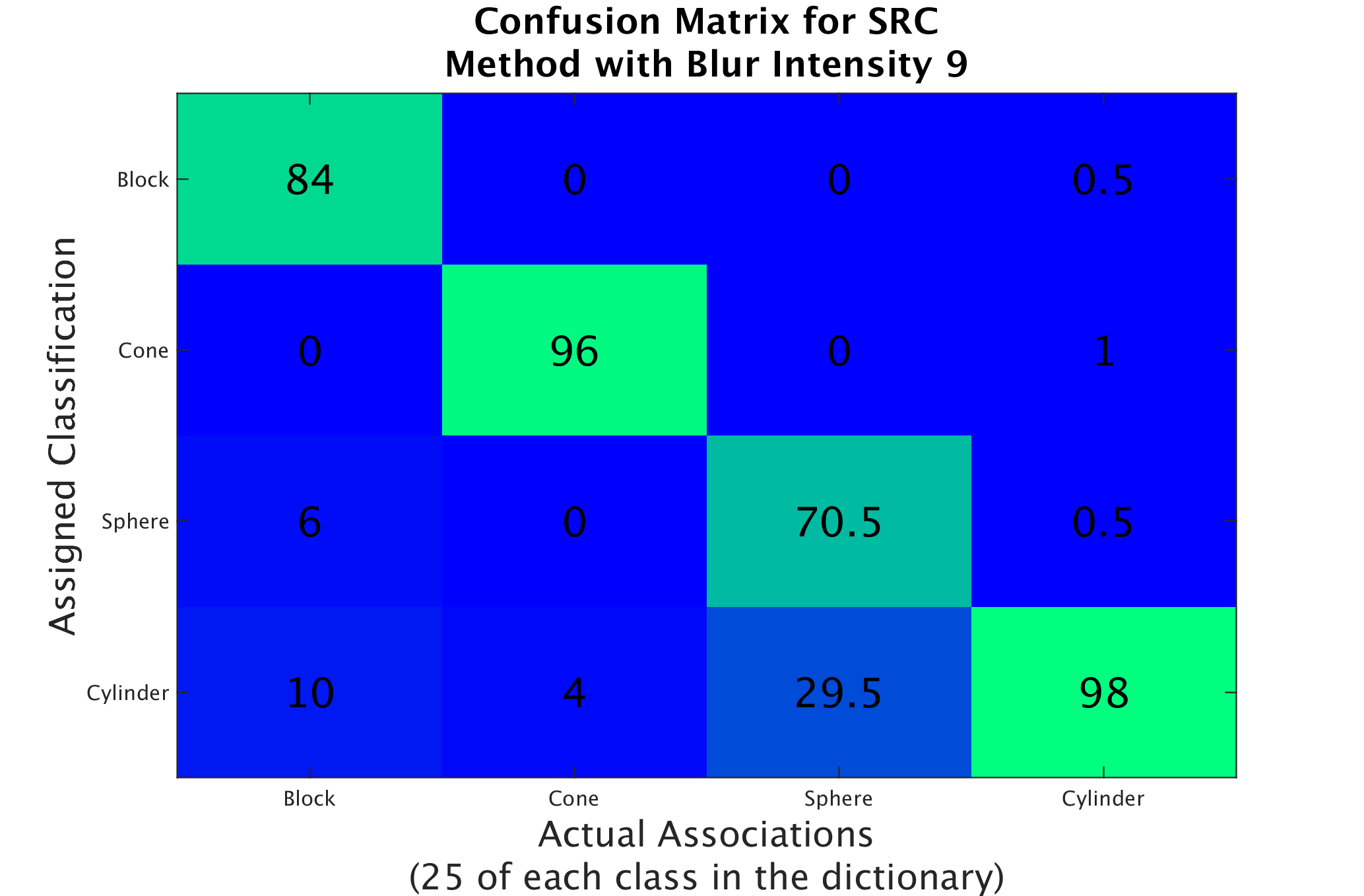}\\\vspace{.1cm}
\includegraphics[width=4.2cm]{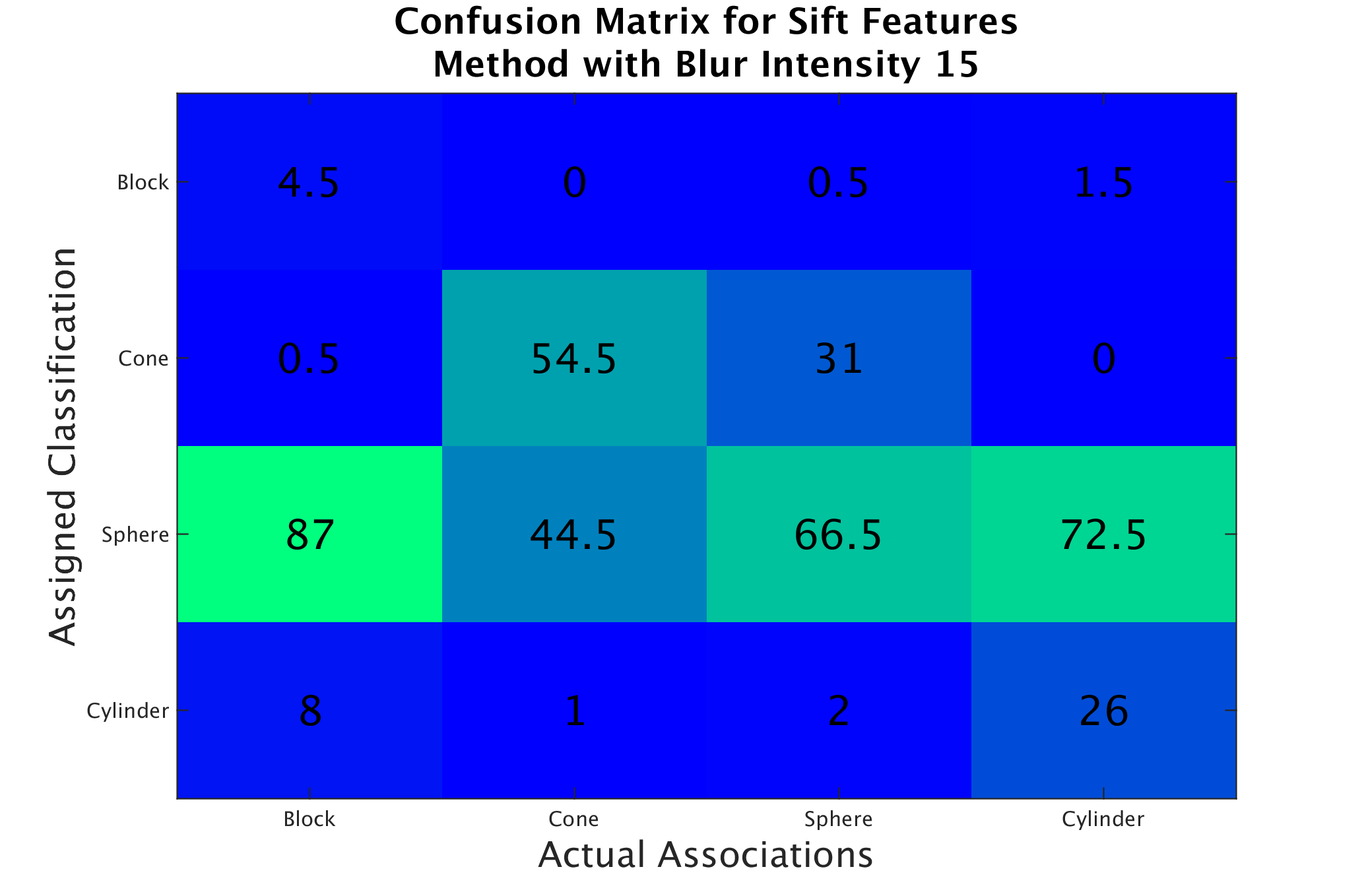}
\includegraphics[width=4.2cm]{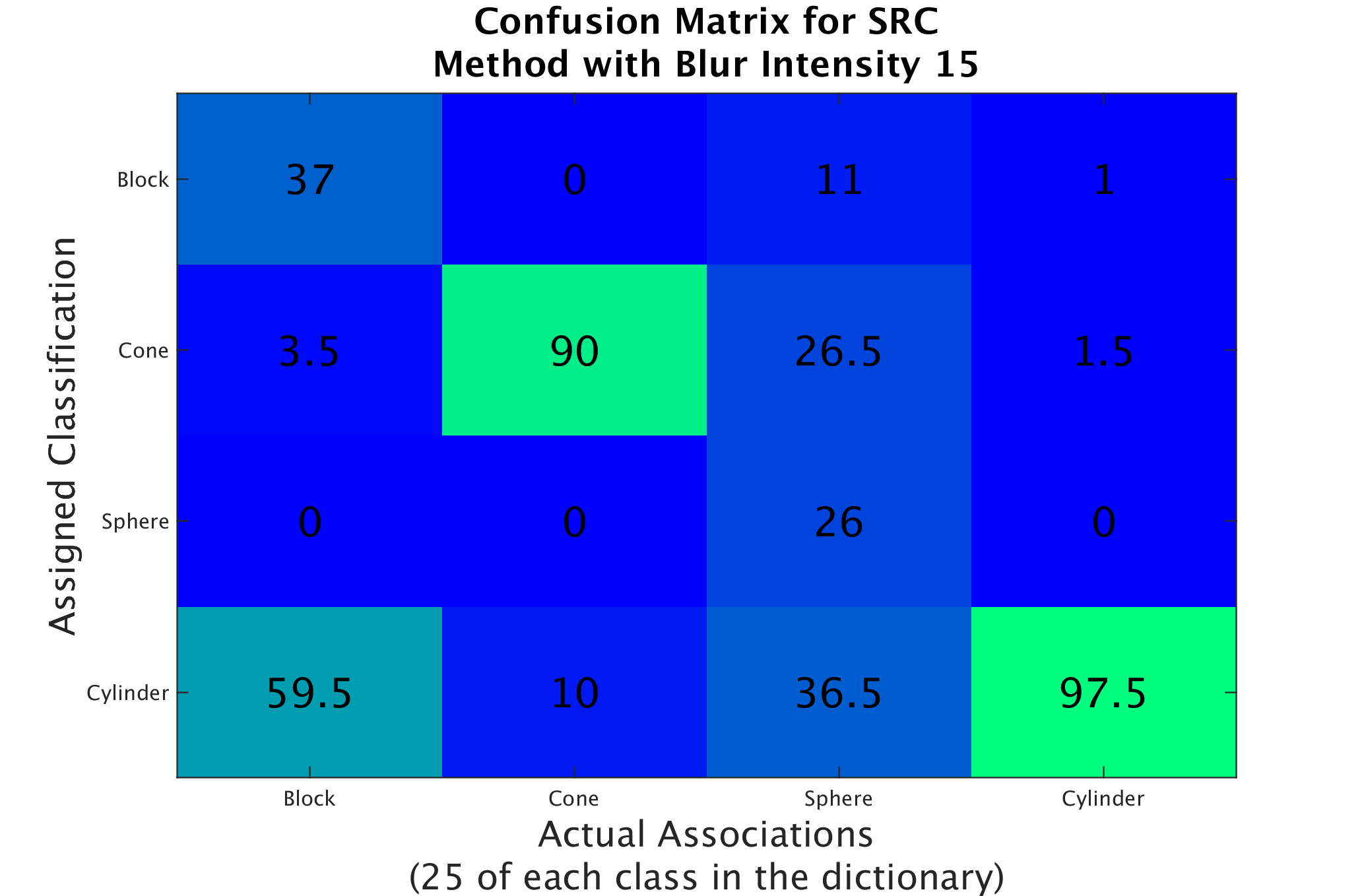}
\caption{Scaled confusion matrices for mid and high level blurring with SRC and SIFT SVM under the equal training.  The columns represent the actual tested image class and the rows their classified class (sums of columns are 100\%).}\label{blurheat}
\end{figure}

\indent The SRC approach was able to sustain better classification rates over higher intensity blur than the popular SIFT SVM method in our tests.  In fact, the SRC approach was able to vastly outperform the SIFT SVM method with increased blurring to a degree greater than with the noise, as Figure \ref{blurcomp} demonstrates.  As we see, as blurring reach a certain severity, the added training became irrelevant for the SIFT SVM approach, whereas the SRC was still able to produce usable classification rates.  With regards to each class, the SRC method showed resiliency in all categories but with spheres, which turned out to be the class that gave the method the most challenge.  The blur did little to impact the classification of cones and cylinders, though under heavy blurring, the SRC approach tended to confuse blocks and spheres with cylinders.  That said, the SIFT SVM approach appeared to fall victim to the rounding type effect heavy blurring had on tests and had difficulty discerning blocks, cones, and cylinders from spheres.  Strategies that rely heavily on the shape of the object, such as SIFT feature-based algorithms, may be more susceptible to these types of errors.  Figure \ref{blurheat} provides the confusion matrices for our blur experiment.

\indent Despite the troubles, given that the SRC approach proved to have an accuracy rate higher than 60\% even in the most intense blurring examples we tried, we see that this strategy may hold a great deal of promise.  In times where SAS imaging fails to produce crisp images, the SRC algorithm can still provide reliable classification rates without the troubles associated with approaches such as the SIFT SVM. 

\subsection{Foreign Object Detection}

In \cite{kriminger2015online}, the authors outlined an active Sonar image classification system that had a built in method to find new objects not found in the training dictionary.  We wished to see how the sparsity concentration index (SCI) outlined in Wright \emph{et al} could perform in a similar task of identifying images outside the training set and, importantly, keeping them from automatically being misclassified into one of the predetermined static number of classes.  The SCI metric for a coefficient vector $\bolx$ found through a SRC approach is given by:
\beqna
\text{SCI}(\bolx)=\frac{\lp \max_{i}\onenorm{\delta_i(\bolx)}/\onenorm{\bolx}\rp k-1}{k-1}
\eeqna
The key to this metric is that it tends to 1 if $\bolx$ is sparse and to 0 otherwise.  Since SRC approaches attempt to force $\bolx$ to be sparse based on a constructed dictionary $A$, if a test vector is foreign with respect to the elements of $A$, then there is a good chance that $\bolx$ we not be very sparse.  Therefore, we can decide to reject a test vector's classification if its SCI value is below a threshold $\kappa\in(0,1)$.  

\indent To test the SCI metric we used the two smaller classes from the RAW SAS dataset, the pipes and toruses which we call the foreign classes, and tested them against dictionaries built with only the main classes.  We varied the dictionary sizes between 25, 35, and 45 members of each main class as to get a better idea of how the training sample can influence this metric.  The tests consisted of 20 remaining elements of the main classes and 10 from both of the foreign classes and we found by the SCI and the assigned classes.  With these, we selected several threshold values $\kappa$ and evaluated how well that value discerned the foreign objects from the main ones as well as how this influenced the overall classification performance of the exercise.

\begin{figure}[t]
\centering
\includegraphics[width=4cm]{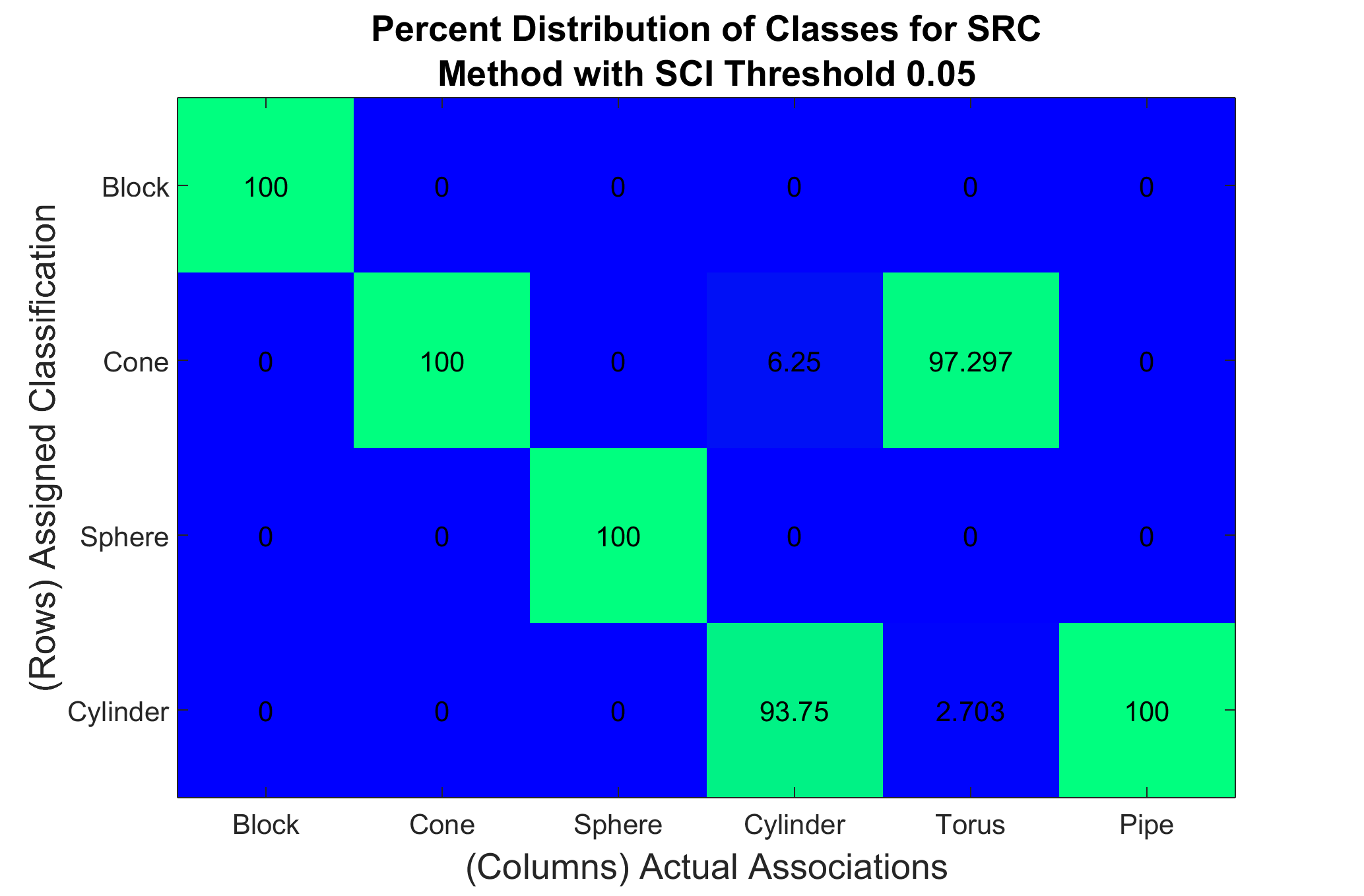}
\includegraphics[width=4cm]{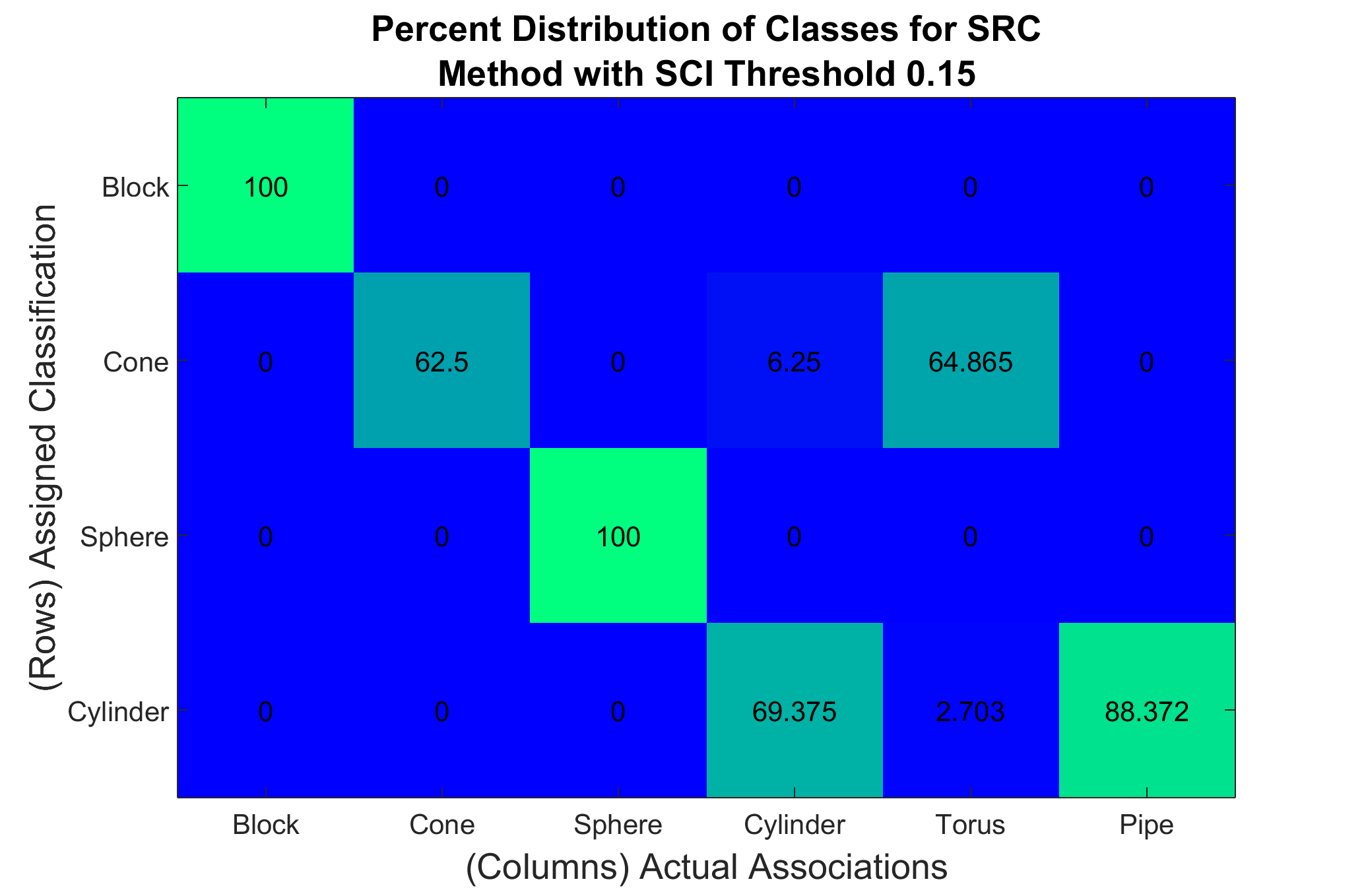}\\
\includegraphics[width=4cm]{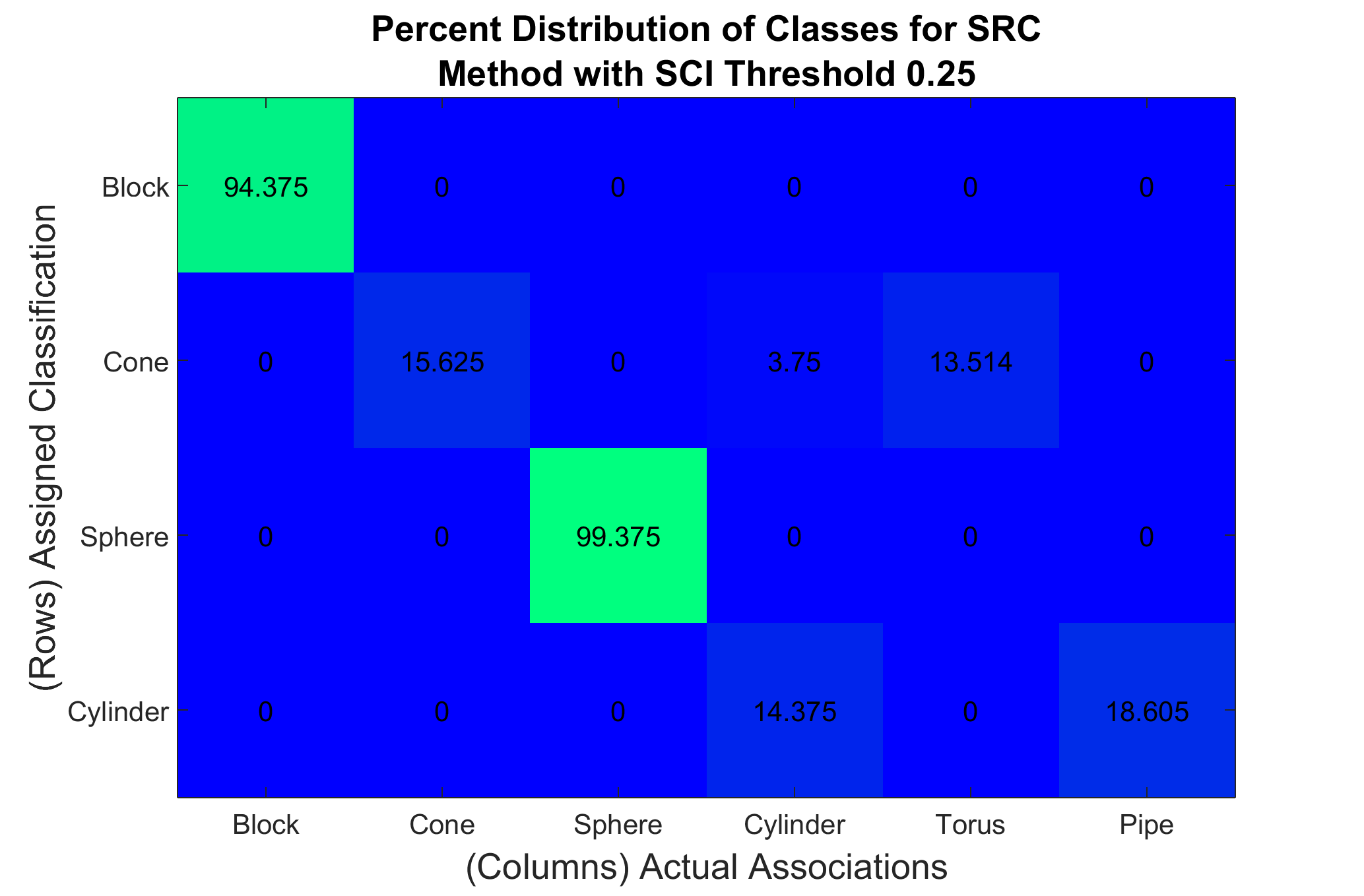}
\caption{Classification rate over different $\kappa$ of dictionaries with 45 elements from each main class.}\label{scikappa}
\end{figure}

\indent We found that the SCI is able to distinguish to some extent between the objects in the dictionary and the newly introduced torus and pipe.  As Table \ref{scis} shows, the SCI metric did its best discriminating between the main and foreign classes when the dictionaries were at their largest.  The lower SCI values for cones and spheres turned out to be an interesting finding and could potentially be due to their smaller are on the target chips, though with larger dictionaries, they were able to differentiate themselves from the foreign classes on average.  When looking at the classification rates, as Figure \ref{scikappa} demonstrates with dictionaries containing 45 elements from each main class, we see some of the problems with using the SCI to weed out foreign objects.  For example, while the pipes had on average a lower SCI value, their variance was less than that of the cones and spheres, so when we increased the threshold $\kappa$ to .15 and .25, we saw many correctly classified cones and spheres removed.  At that same time, even with the highest thresholds, the blocks and cylinders were barely effected.  Therefore, the SCI metric may be useful in certain cases such as with our dataset if we were concerned with finding only blocks and cylinders, but more investigation must be done to refine the SCI so it can become a tool in foreign object detection.

\begin{table}[t]
\centering\normalsize
\begin{tabular}{c||c|c|c}
\multirow{2}{*}{\textbf{Object}} & \multicolumn{3}{c}{\textbf{SCI with Training Sizes of:}}\\
& 25 & 35 & 45\\\hline\hline
Block & .478 (.130) & .463 (.106) & .683 (.151)\\
Cone & .169 (.059) & .175 (.051) & .471 (.154)\\
Cylinders & .501 (.088) & .438 (.079) &  .745 (.089)\\
Spheres & .184 (.062) & .181 (.056) & .466 (.182)\\\hline
Torus & .173 (.038) & .201 (.033) & .195 (.053)\\
Pipe & .193 (.054) & .216 (.046) & .329 (.097)\\\hline
\end{tabular}
\caption{Average SCI values with standard deviations for our tests on dictionaries that used 25, 35, and then 45 elements from each main class.}\label{scis}
\end{table}

\section{Discussions and Further Work}

In this paper we have applied the SRC methodology to Sonar ATR and found robust performance in varying amounts of noise and blur. Furthermore this method continues to exhibit superior performance to the popular SIFT SVM based approach in limited training size regimes. It is well known however that the performance of SRC is sensitive to the correctness of pose estimation in both the training and test samples. Given this an important next step would be to examine the problem if image pose.

\indent The peculiarities of the Sonar imaging modality, however, demands solutions that are somewhat different from those that are successful for optical images (in particular for facial images). In our experiments we carefully focused on Sonar images obtained from a normal side-scan capture. More generally, the angles at which an object faces the Sonar device greatly influences its appearance inasmuch as a slight adjustment in its position can render vastly different image projections. Given this we posit that it would be prudent to focus on translation invariant methods when extracting image features conditioned on the pose of the object under consideration. This translational invariance property would also enable the classification of images with inexact window sizes. Our future investigations aim to resolve these issues while also incorporating powerful Bayesian priors on the underlying dictionary structure \cite{srinivas2015structured} \cite{srinivas2012IGT} \cite{raj2014hbmap} to enable robust classification performance in increasingly difficult clutter environments.

\bibliographystyle{IEEEtran}
\bibliography{refOceans2015}



%
%
%

\end{document}